\documentclass{article} 
\pdfoutput=1
\usepackage{iclr2017_conference,times}
\usepackage{url}
\usepackage{multirow}
\usepackage{graphicx}
\usepackage{mathtools}
\usepackage{subcaption}
\usepackage{caption}
\usepackage{array}
\usepackage{wrapfig,lipsum,booktabs}
\usepackage{color}
\usepackage{amssymb,mathtools}
\usepackage[T1]{fontenc}
\usepackage[utf8]{inputenc}
\usepackage{authblk}

\title{\Large Outrageously Large Neural Networks: \\
The Sparsely-Gated Mixture-of-Experts Layer}

\author[1]{Noam Shazeer}
\author[1]{Azalia Mirhoseini\thanks{Equally major contributors} \thanks{Work done as a member of the Google Brain Residency program (g.co/brainresidency)}~}
\author[2]{Krzysztof Maziarz$^*$}
\author[1]{Andy Davis}
\author[1]{Quoc Le}
\author[1]{Geoffrey Hinton}
\author[1]{Jeff Dean}
\affil[1]{Google Brain, \{noam,azalia,andydavis,qvl,geoffhinton,jeff\}@google.com}
\affil[2]{Jagiellonian University, Cracow, krzysztof.maziarz@student.uj.edu.pl}

\begin{document}

\maketitle

\begin{abstract}

The capacity of a neural network to absorb information is limited by its number of parameters.  Conditional computation, where parts of the network are active on a per-example basis, has been proposed in theory as a way of dramatically increasing model capacity without a proportional increase in computation.  In practice, however, there are significant algorithmic and performance challenges.  In this work, we address these challenges and finally realize the promise of conditional computation, achieving greater than 1000x improvements in model capacity with only minor losses in computational efficiency on modern GPU clusters.  We introduce a Sparsely-Gated Mixture-of-Experts layer (MoE), consisting of up to thousands of feed-forward sub-networks.  A trainable gating network determines a sparse combination of these experts to use for each example.  We apply the MoE to the tasks of language modeling and machine translation, where model capacity is critical for absorbing the vast quantities of knowledge available in the training corpora.  We present model architectures in which a MoE with up to 137 billion parameters is applied convolutionally between stacked LSTM layers.  On large language modeling and machine translation benchmarks, these models achieve significantly better results than state-of-the-art at lower computational cost.

\end{abstract}

\section{Introduction and Related Work}
\subsection{Conditional Computation}


Exploiting scale in both training data and model size has been central to the success of deep learning. When datasets are sufficiently large, increasing the capacity (number of parameters) of neural networks can give much better prediction accuracy.  This has been shown in domains such as text \citep{sutskever2014sequence,bahdanau2014neural,RafalNoam16,GNMT}, images \citep{Imagenet,qvl2012building}, and audio \citep{hinton2012deep,DeepSpeech2}.   For typical deep learning models, where the entire model is activated for every example, this leads to a roughly quadratic blow-up in training costs, as both the model size and the number of training examples increase.  Unfortunately, the advances in computing power and distributed computation fall short of meeting such demand. 

Various forms of conditional computation have been proposed as a way to increase model capacity without a proportional increase in computational costs \citep{Davis13:CondComp, Bengio13:CondComp, eigen2013learning, Denoyer14:CondComp,  Cho14, Bengio15:CondComp, Almahairi15}.  In these schemes, large parts of a network are active or inactive on a per-example basis.  The gating decisions may be binary or sparse and continuous, stochastic or deterministic.  Various forms of reinforcement learning and back-propagation are proposed for trarining the gating decisions. 

While these ideas are promising in theory, no work to date has yet demonstrated massive improvements in model capacity, training time, or model quality.  We blame this on a combination of the following challenges:

\begin{itemize}
  \item Modern computing devices, especially GPUs, are much faster at arithmetic than at branching.   Most of the works above recognize this and propose turning on/off large chunks of the network with each gating decision.
  \item Large batch sizes are critical for performance, as they amortize the costs of parameter transfers and updates.  Conditional computation reduces the batch sizes for the conditionally active chunks of the network.
  \item Network bandwidth can be a bottleneck.  A cluster of GPUs may have computational power thousands of times greater than the aggregate inter-device network bandwidth.  To be computationally efficient, the relative computational versus network demands of an algorithm must exceed this ratio.   Embedding layers, which can be seen as a form of conditional computation, are handicapped by this very problem.  Since the embeddings generally need to be sent across the network, the number of (example, parameter) interactions is limited by network bandwidth instead of computational capacity.
  \item Depending on the scheme, loss terms may be necessary to achieve the desired level of sparsity per-chunk and/or per example. \cite{Bengio15:CondComp} use three such terms.  These issues can affect both model quality and load-balancing.
  \item Model capacity is most critical for very large data sets.  The existing literature on conditional computation deals with relatively small image recognition data sets consisting of up to 600,000 images.  It is hard to imagine that the labels of these images provide a sufficient signal to adequately train a model with millions, let alone billions of parameters.
\end{itemize}
 
 In this work, we for the first time address all of the above challenges and finally realize the promise of conditional computation. We obtain greater than 1000x improvements in model capacity with only minor losses in computational efficiency and significantly advance the state-of-the-art results on public language modeling and translation data sets.
 
\subsection{Our Approach: The Sparsely-Gated Mixture-of-Experts Layer}

\begin{figure}
    \includegraphics[width=.90\textwidth]{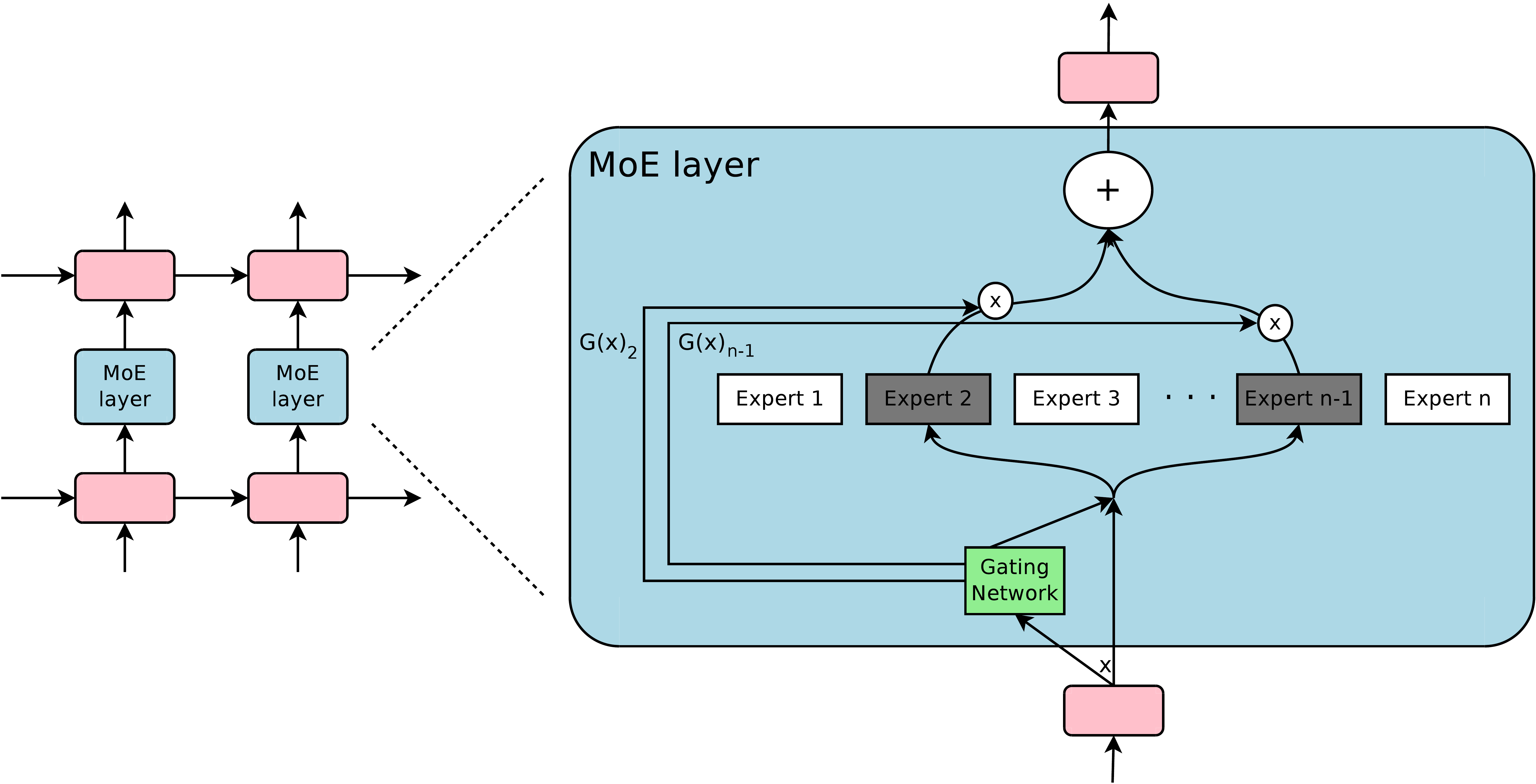}
    \caption{A Mixture of Experts (MoE) layer embedded within a recurrent language model. In this case, the sparse gating function selects two experts to perform computations. Their outputs are modulated by the outputs of the gating network.}
    \label{fig:moe}
\end{figure}

Our approach to conditional computation is to introduce a new type of general purpose neural network component: a Sparsely-Gated Mixture-of-Experts Layer (MoE).  The MoE consists of a number of experts, each a simple feed-forward neural network, and a trainable gating network which selects a sparse combination of the experts to process each input (see Figure \ref{fig:moe}).  All parts of the network are trained jointly by back-propagation.

While the introduced technique is generic, in this paper we focus on language modeling and machine translation tasks, which are known to benefit from very large models. In particular, we apply a MoE convolutionally between stacked LSTM layers \citep{Hochreiter:1997:LSM}, as in Figure \ref{fig:moe}.  The MoE is called once for each position in the text, selecting a potentially different combination of experts at each position.  The different experts tend to become highly specialized based on syntax and semantics (see Appendix \ref{sec:appendixmt} Table \ref{tab:experts}).  On both language modeling and machine translation benchmarks, we improve on best published results at a fraction of the computational cost.

\subsection{Related work on Mixtures of Experts}

Since its introduction more than two decades ago \citep{Jacobs91Adaptive,Jordan1994HME}, the mixture-of-experts approach has been the subject of much research.  Different types of expert architectures hae been proposed such as SVMs \citep{Collobert02PMS}, Gaussian Processes \citep{Tresp2001Mixture,Theis2015Generative,Deisenroth15Distributed}, Dirichlet Processes \citep{Shahbaba09NMU}, and deep networks.  Other work has focused on different expert configurations such as a hierarchical structure \citep{Yao09Hierarchical}, infinite numbers of experts \citep{Rasmussen02Infinite}, and adding experts sequentially \citep{Aljundi16}.  \cite{Garmash2016ensemble} suggest an ensemble model in the format of mixture of experts for machine translation. The gating network is trained on a pre-trained ensemble NMT model.

The works above concern top-level mixtures of experts.  The mixture of experts is the whole model.   \cite{eigen2013learning} introduce the idea of using multiple MoEs with their own gating networks as parts of a deep model.  It is intuitive that the latter approach is more powerful, since complex problems may contain many sub-problems each requiring different experts.  They also allude in their conclusion to the potential to introduce sparsity, turning MoEs into a vehicle for computational computation.  

Our work builds on this use of MoEs as a general purpose neural network component.  While \cite{eigen2013learning} uses two stacked MoEs allowing for two sets of gating decisions, our convolutional application of the MoE allows for different gating decisions at each position in the text.   We also realize sparse gating and demonstrate its use as a practical way to massively increase model capacity.

\section{The Structure of the Mixture-of-Experts layer}\label{sec:gating}

The Mixture-of-Experts (MoE) layer consists of a set of $n$ ``expert networks" $E_1, \cdots, E_n$, and a ``gating network" $G$ whose output is a sparse $n$-dimensional vector.  Figure \ref{fig:moe} shows an overview of the MoE module. The experts are themselves neural networks, each with their own parameters.  Although in principle we only require that the experts accept the same sized inputs and produce the same-sized outputs, in our initial investigations in this paper, we restrict ourselves to the case where the models are feed-forward networks with identical architectures, but with separate parameters.

Let us denote by $G(x)$ and $E_i(x)$ the output of the gating network and the output of the $i$-th expert network for a given input $x$. The output $y$ of the MoE module can be written as follows:

\begin{equation}
y = \sum_{i=1}^{n}G(x)_iE_i(x)
\end{equation}

We save computation based on the sparsity of the output of $G(x)$.  Wherever $G(x)_i=0$, we need not compute $E_i(x)$.  In our experiments, we have up to thousands of experts, but only need to evaluate a handful of them for every example. If the number of experts is very large, we can reduce the branching factor by using a two-level hierarchical MoE. In a hierarchical MoE, a primary gating network chooses a sparse weighted combination of ``experts", each of which is itself a secondary mixture-of-experts with its own gating network. In the following we focus on ordinary MoEs. We provide more details on hierarchical MoEs in Appendix \ref{sec:hierarchical}.

Our implementation is related to other models of conditional computation. A MoE whose experts are simple weight matrices is similar to the parameterized weight matrix proposed in \citep{Cho14}.  A MoE whose experts have one hidden layer is similar to the block-wise dropout described in \citep{Bengio15:CondComp}, where the dropped-out layer is sandwiched between fully-activated layers.

\subsection{Gating Network}

\paragraph{Softmax Gating:} A simple choice of non-sparse gating function \citep{Jordan1994HME} is to multiply the input by a trainable weight matrix $W_g$ and then apply the $Softmax$ function.

\begin{equation}\label{eq:softmax}
G_\sigma(x) = Softmax(x \cdot W_g)
\end{equation}

\paragraph{Noisy Top-K Gating:}\label{sec:noisytopk} We add two components to the Softmax gating network: sparsity and noise.  Before taking the softmax function, we add tunable Gaussian noise, then keep only the top k values, setting the rest to $-\infty$ (which causes the corresponding gate values to equal $0$).  The sparsity serves to save computation, as described above.  While this form of sparsity creates some theoretically scary discontinuities in the output of gating function, we have not yet observed this to be a problem in practice.  The noise term helps with load balancing, as will be discussed in Appendix \ref{sec:load}.  The amount of noise per component is controlled by a second trainable weight matrix $W_{noise}$.

\begin{equation}\label{eq:g}
G(x) = Softmax(KeepTopK(H(x), k))
\end{equation}

\begin{equation}\label{eq:noise}
H(x)_i = (x \cdot W_g)_i + StandardNormal() \cdot Softplus((x \cdot W_{noise})_i)
\end{equation}

\begin{equation}\label{eq:keeptopk}
KeepTopK(v, k)_i = \begin{cases}
            v_i & \text{if $v_i$ is in the top $k$ elements of $v$.} \\
            -\infty & \text{otherwise.}
        \end{cases}
\end{equation}

\paragraph{Training the Gating Network}
We train the gating network by simple back-propagation, along with the rest of the model.  If we choose $k>1$, the gate values for the top k experts have nonzero derivatives with respect to the weights of the gating network.  This type of occasionally-sensitive behavior is described in \citep{Bengio13:CondComp} with respect to noisy rectifiers.  Gradients also back-propagate through the gating network to its inputs.   Our method differs here from \citep{Bengio15:CondComp} who use boolean gates and a REINFORCE-style approach to train the gating network.

\section{Addressing Performance Challenges}
\label{sec:performance}
\subsection{The Shrinking Batch Problem}
On modern CPUs and GPUs, large batch sizes are necessary for computational efficiency, so as to amortize the overhead of parameter loads and updates.  If the gating network chooses $k$ out of $n$ experts for each example, then for a batch of $b$ examples, each expert receives a much smaller batch of approximately $\frac{kb}{n}\ll b$ examples.  This causes a naive MoE implementation to become very inefficient as the number of experts increases.  The solution to this shrinking batch problem is to make the original batch size as large as possible.  However, batch size tends to be limited by the memory necessary to store activations between the forwards and backwards passes.  We propose the following techniques for increasing the batch size:

\paragraph{Mixing Data Parallelism and Model Parallelism:}  In a conventional distributed training setting, multiple copies of the model on different devices asynchronously process distinct batches of data, and parameters are synchronized through a set of parameter servers.  In our technique, these different batches run synchronously so that they can be combined for the MoE layer.  We distribute the standard layers of the model and the gating network according to conventional data-parallel schemes, but keep only one shared copy of each expert.  Each expert in the MoE layer receives a combined batch consisting of the relevant examples from all of the data-parallel input batches.   The same set of devices function as data-parallel replicas (for the standard layers and the gating networks) and as model-parallel shards (each hosting a subset of the experts).  If the model is distributed over $d$ devices, and each device processes a batch of size $b$, each expert receives a batch of approximately $\frac{kbd}{n}$ examples. Thus, we achieve a factor of $d$ improvement in expert batch size. 

In the case of a hierarchical MoE (Section \ref{sec:hierarchical}), the primary gating network employs data parallelism, and the secondary MoEs employ model parallelism.  Each secondary MoE resides on one device.

This technique allows us to increase the number of experts (and hence the number of parameters) by proportionally increasing the number of devices in the training cluster.   The total batch size increases, keeping the batch size per expert constant.   The memory and bandwidth requirements per device also remain constant, as do the step times, as does the amount of time necessary to process a number of training examples equal to the number of parameters in the model.  It is our goal to train a trillion-parameter model on a trillion-word corpus.  We have not scaled our systems this far as of the writing of this paper, but it should be possible by adding more hardware.

\paragraph{Taking Advantage of Convolutionality:} In our language models, we apply the same MoE to each time step of the previous layer. If we wait for the previous layer to finish, we can apply the MoE to all the time steps together as one big batch.  Doing so increases the size of the input batch to the MoE layer by a factor of the number of unrolled time steps.

\paragraph{Increasing Batch Size for a Recurrent MoE:}  We suspect that even more powerful models may involve applying a MoE recurrently.  For example, the weight matrices of a LSTM or other RNN could be replaced by a MoE.   Sadly, such models break the convolutional trick from the last paragraph, since the input to the MoE at one timestep depends on the output of the MoE at the previous timestep.  \cite{Gruslys16} describe a technique for drastically reducing the number of stored activations in an unrolled RNN, at the cost of recomputing forward activations.   This would allow for a large increase in batch size.

\subsection{Network Bandwidth}
Another major performance concern in distributed computing is network bandwidth.  Since the experts are stationary (see above) and the number of gating parameters is small, most of the communication involves sending the inputs and outputs of the experts across the network.  To maintain computational efficiency, the ratio of an expert's computation to the size of its input and output must exceed the ratio of computational to network capacity of the computing device.   For GPUs, this may be thousands to one.  In our experiments, we use experts with one hidden layer containing thousands of RELU-activated units.  Since the weight matrices in the expert have sizes $input$\_${size} \times hidden$\_${size}$ and $hidden$\_${size} \times output$\_${size}$, the ratio of computation to input and output is equal to the size of the hidden layer.  Conveniently, we can increase computational efficiency simply by using a larger hidden layer, or more hidden layers.

\section{Balancing Expert Utilization}  
\label{sec:losses}

We have observed that the gating network tends to converge to a state where it always produces large weights for the same few experts.  This imbalance is self-reinforcing, as the favored experts are trained more rapidly and thus are selected even more by the gating network.   \cite{eigen2013learning} describe the same phenomenon, and use a hard constraint at the beginning of training to avoid this local minimum.  \cite{Bengio15:CondComp} include a soft constraint on the batch-wise average of each gate.\footnote{\cite{Bengio15:CondComp} also include two additional losses.  One controls per-example sparsity, which we do not need since it is enforced by the fixed value of $k$.  A third loss encourages diversity of gate values.  In our experiments, we find that the gate values naturally diversify as the experts specialize (in a virtuous cycle), and we do not need to enforce diversity of gate values.}

We take a soft constraint approach.  We define the importance of an expert relative to a batch of training examples to be the batchwise sum of the gate values for that expert.  We define an additional loss $L_{importance}$, which is added to the overall loss function for the model.  This loss is equal to the square of the coefficient of variation of the set of importance values, multiplied by a hand-tuned scaling factor $w_{importance}$.  This additional loss encourages all experts to have equal importance.

\begin{equation}\label{eq:gateloss}
Importance(X) = \sum_{x \in X}G(x)
\end{equation}

\begin{equation}\label{eq:importanceloss}
L_{importance}(X) = w_{importance} \cdot CV(Importance(X))^2
\end{equation}

While this loss function can ensure equal importance, experts may still receive very different numbers of examples.  For example, one expert may receive a few examples with large weights, and another may receive many examples with small weights.  This can cause memory and performance problems on distributed hardware.  To solve this problem, we introduce a second loss function, $L_{load}$ , which ensures balanced loads.  Appendix \ref{sec:load} contains the definition of this function, along with experimental results.

\section{Experiments}

\subsection{1 Billion Word Language Modeling Benchmark}\label{sec:lm}

\paragraph{Dataset:} This dataset, introduced by \citep{chelba2013one} consists of shuffled unique sentences from news articles, totaling approximately 829 million words, with a vocabulary of 793,471 words.

\begin{figure}[h!]
\centering
\begin{minipage}{.49\textwidth}
  \centering
  \includegraphics[width=1.0\linewidth]{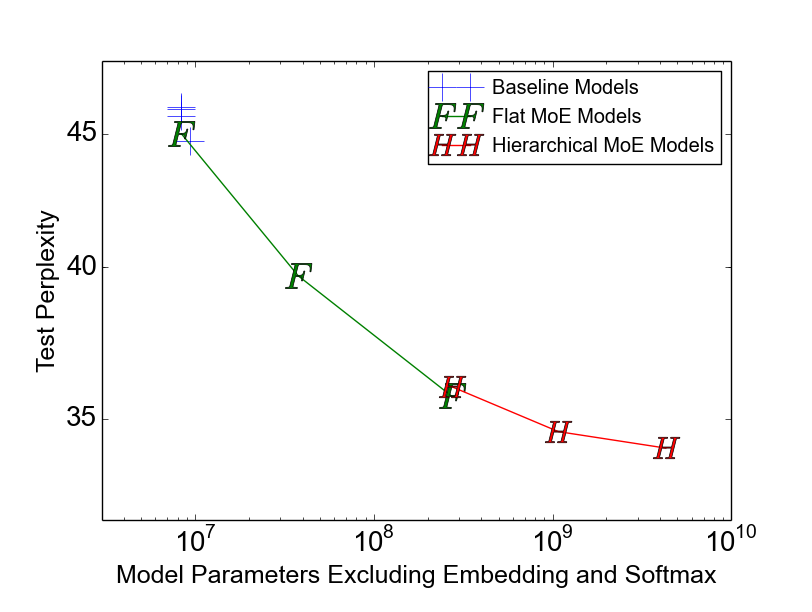}
\end{minipage}
\begin{minipage}{.49\textwidth}
  \centering
  \includegraphics[width=1.0\linewidth]{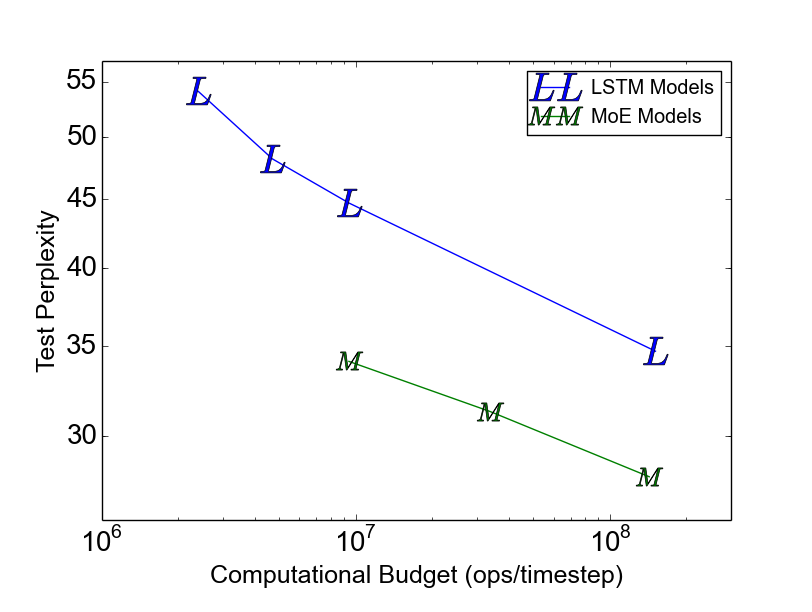}
\end{minipage}
\caption{Model comparison on 1-Billion-Word Language-Modeling Benchmark.  On the left, we plot test perplexity as a function of model capacity for models with similar computational budgets of approximately 8-million-ops-per-timestep.  On the right, we plot test perplexity as a function of computational budget.  The top line represents the LSTM models from \citep{RafalNoam16}.  The bottom line represents 4-billion parameter MoE models with different computational budgets.}
\label{fig:lm1b}
\end{figure}

\paragraph{Previous State-of-the-Art:} The best previously published results \citep{RafalNoam16} use models consisting of one or more stacked Long Short-Term Memory (LSTM) layers \citep{Hochreiter:1997:LSM,Gers:2000:LFC}.  The number of parameters in the LSTM layers of these models vary from 2 million to 151 million.  Quality increases greatly with parameter count, as do computational costs.  Results for these models form the top line of Figure \ref{fig:lm1b}-right.

\paragraph{MoE Models:} Our models consist of two stacked LSTM layers with a MoE layer between them (see Figure \ref{fig:moe}).  We vary the sizes of the layers and the number of experts.   For full details on model architecture, training regimen, additional baselines and results, see Appendix \ref{sec:appendixlm1b}.   

\paragraph{Low Computation, Varied Capacity:} To investigate the effects of adding capacity, we trained a series of MoE models all with roughly equal computational costs: about 8 million multiply-and-adds per training example per timestep in the forwards pass, excluding the softmax layer.  We call this metric (ops/timestep).  We trained models with flat MoEs containing 4, 32, and 256 experts, and models with hierarchical MoEs containing 256, 1024, and 4096 experts.  Each expert had about 1 million parameters.  For all the MoE layers, 4 experts were active per input.

The results of these models are shown in Figure \ref{fig:lm1b}-left.   The model with 4 always-active experts performed (unsurprisingly) similarly to the computationally-matched baseline models, while the largest of the models (4096 experts) achieved an impressive 24\% lower perplexity on the test set.

\begin{table}[!htbp]
\caption{ Summary of high-capacity MoE-augmented models with varying computational budgets, vs. best previously published results \citep{RafalNoam16}.  Details in Appendix \ref{sec:appendixlm1b}.}
\label{tab:lm1bshort}
\begin{center}
\setlength\tabcolsep{3pt}
\scalebox{0.8}{
\begin{tabular}{l|c|c|c|c|c|c}
\hline \hline
 & Test        & Test       & \#Parameters         & ops/timestep & Training & TFLOPS\\
 & Perplexity  & Perplexity & excluding embedding  &     & Time & /GPU\\
 & 10 epochs & 100 epochs & and softmax layers &  & 10 epochs & \\
\hline
Best Published Results & 34.7 & 30.6 & 151 million & 151 million & 59 hours, 32 k40s & 1.09\\
\hline
Low-Budget MoE Model & 34.1 & & 4303 million & 8.9 million & 15 hours, 16 k40s & 0.74\\
Medium-Budget MoE Model & 31.3 & & 4313 million & 33.8 million & 17 hours, 32 k40s & 1.22\\
High-Budget MoE Model & \textbf{28.0} & & 4371 million & 142.7 million & 47 hours, 32 k40s & \textbf{1.56}\\
\hline \hline
\end{tabular} 
}
\end{center}
\end{table}

\paragraph{Varied Computation, High Capacity:}  In addition to the largest model from the previous section, we trained two more MoE models with similarly high capacity (4 billion parameters), but higher computation budgets.  These models had larger LSTMs, and fewer but larger and experts.  Details can be found in Appendix \ref{sec:expensive}.  Results of these three models form the bottom line of Figure \ref{fig:lm1b}-right.  Table \ref{tab:lm1bshort} compares the results of these models to the best previously-published result on this dataset .  Even the fastest of these models beats the best published result (when controlling for the number of training epochs), despite requiring only 6\% of the computation.  

\paragraph{Computational Efficiency:} We trained our models using TensorFlow \citep{Abadi16} on clusters containing 16-32 Tesla K40 GPUs.  For each of our models, we determine computational efficiency in TFLOPS/GPU by dividing the number of floating point operations required to process one training batch by the observed step time and the number of GPUs in the cluster.  The operation counts used here are higher than the ones we report in our ops/timestep numbers in that we include the backwards pass, we include the importance-sampling-based training of the softmax layer, and we count a multiply-and-add as two separate operations.  For all of our MoE models, the floating point operations involved in the experts represent between 37\% and 46\% of the total. 

For our baseline models wtih no MoE, observed computational efficiency ranged from 1.07-1.29 TFLOPS/GPU.  For our low-computation MoE models, computation efficiency ranged from 0.74-0.90 TFLOPS/GPU, except for the 4-expert model which did not make full use of the available parallelism.  Our highest-computation MoE model was more efficient at 1.56 TFLOPS/GPU, likely due to the larger matrices.   These numbers represent a significant fraction of the theoretical maximum of 4.29 TFLOPS/GPU claimed by NVIDIA.  Detailed results are in Appendix \ref{sec:appendixlm1b}, Table \ref{tab:lm1bresults}.

\subsection{100 Billion Word Google News Corpus}

\begin{figure}[h!]
\centering
\includegraphics[width=0.5\linewidth]{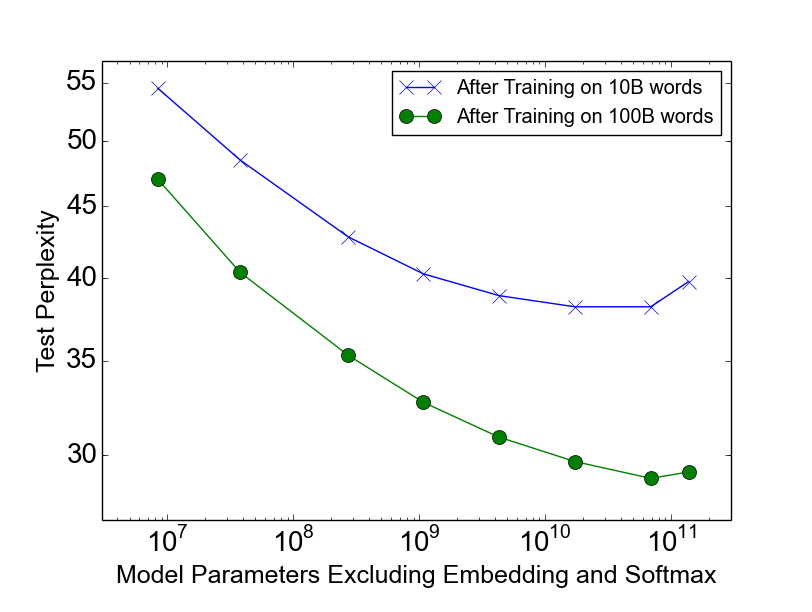}
\caption{Language modeling on a 100 billion word corpus.  Models have similar computational budgets (8 million ops/timestep).}
\label{fig:gn11}
\end{figure}

On the 1-billion-word corpus, adding additional capacity seems to produce diminishing returns as the number of parameters in the MoE layer exceeds 1 billion, as can be seen in Figure \ref{fig:lm1b}-left.   We hypothesized that for a larger training set, even higher capacities would produce significant quality improvements.

We constructed a similar training set consisting of shuffled unique sentences from Google's internal news corpus, totalling roughly 100 billion words.  Similarly to the previous section, we tested a series of models with similar computational costs of about 8 million ops/timestep.  In addition to a baseline LSTM model, we trained models augmented with MoE layers containing 32, 256, 1024, 4096, 16384, 65536, and 131072 experts.  This corresponds to up to 137 billion parameters in the MoE layer.  Details on architecture, training, and results are given in Appendix \ref{sec:appendixgn11}. 

\paragraph{Results:} Figure \ref{fig:gn11} shows test perplexity as a function of capacity after training on 10 billion words (top line) and 100 billion words (bottom line).  When training over the full 100 billion words, test perplexity improves significantly up to 65536 experts (68 billion parameters), dropping 39\% lower than the computationally matched baseline, but degrades at 131072 experts, possibly a result of too much sparsity.  The widening gap between the two lines demonstrates (unsurprisingly) that increased model capacity helps more on larger training sets.

Even at 65536 experts (99.994\% layer sparsity), computational efficiency for the model stays at a respectable 0.72 TFLOPS/GPU.

\subsection{Machine Translation (Single Language Pair)}
\label{sec:mt}
\paragraph{Model Architecture:} Our model was a modified version of the GNMT model described in~\citep{GNMT}.  To reduce computation, we decreased the number of LSTM layers in the encoder and decoder from 9 and 8 to 3 and 2 respectively.  We inserted MoE layers in both the encoder (between layers 2 and 3) and the decoder (between layers 1 and 2).  Each MoE layer contained up to 2048 experts each with about two million parameters, adding a total of about 8 billion parameters to the models.  Further details on model architecture, testing procedure and results can be found in Appendix \ref{sec:appendixmt}.

\paragraph{Datasets:} We benchmarked our method on the WMT'14 En$\rightarrow$Fr and En$\rightarrow$De corpora, whose training sets have 36M sentence pairs and 5M sentence pairs, respectively. The experimental protocols were also similar to those in~\citep{GNMT}: newstest2014 was used as the test set to compare against previous work \citep{LuongPM:2015:EAANMT,Zhou:2016:DeppAtt,GNMT}, while the combination of newstest2012 and newstest2013 was used as the development set.  We also tested the same model on a Google's Production English to French data.

\begin{table}[h!]
\caption{Results on WMT'14 En$\rightarrow$ Fr newstest2014 (bold values represent best results).}
\label{tab:wmtenfr}
\begin{center}
\setlength\tabcolsep{2pt}
\vspace{-5pt}
\scalebox{0.8}{
\begin{tabular}{l|c|c|c|c|c}
\hline \hline
Model & Test & Test & ops/timenstep & Total & Training \\
       & Perplexity & BLEU &  &\#Parameters & Time \\\hline
MoE with 2048 Experts  & 2.69 & 40.35  & 85M & 8.7B & 3 days/64 k40s \\
MoE with 2048 Experts (longer training)  & \textbf{ 2.63} & \textbf{40.56}  & 85M & 8.7B & 6 days/64 k40s \\
\hline
GNMT \citep{GNMT} & 2.79 & 39.22 &  214M & 278M  & 6 days/96 k80s\\
GNMT+RL \citep{GNMT} & 2.96 &39.92 &   214M & 278M & 6 days/96 k80s \\
PBMT \citep{PBMT}  && 37.0&&\\
LSTM (6-layer) \citep{Minh-Thang15} &  & 31.5   &&   \\
LSTM (6-layer+PosUnk) \citep{Minh-Thang15} &  & 33.1 &&   \\
DeepAtt \citep{Zhou:2016:DeppAtt}  &  & 37.7 &&   \\
DeepAtt+PosUnk \citep{Zhou:2016:DeppAtt} & & 39.2  &&   \\
\hline \hline
\end{tabular}
}
\end{center}
\end{table}

\begin{table}[h!]
\caption{Results on WMT'14 En $\rightarrow$ De newstest2014 (bold values represent best results).}
\label{tab:wmtende}
\begin{center}
\vspace{-5pt}
\scalebox{0.8}{
\begin{tabular}{l|c|c|c|c|c}
\hline \hline
Model & Test & Test & ops/timestep &  Total  & Training \\
       & Perplexity & BLEU &   & \#Parameters & Time \\\hline
MoE with 2048 Experts  & \textbf{4.64}  & \textbf{26.03} & 85M & 8.7B & 1 day/64 k40s \\
\hline
GNMT \citep{GNMT} &5.25 & 24.91 & 214M & 278M  & 1 day/96 k80s\\
GNMT +RL \citep{GNMT} &8.08 & 24.66 &  214M & 278M & 1 day/96 k80s\\ 
PBMT \citep{PBMT}  && 20.7 &&\\
DeepAtt \citep{Zhou:2016:DeppAtt}  &  & 20.6 &&  \\
\hline \hline
\end{tabular}
}
\end{center}
\end{table}

\begin{table}[h!]
\caption{Results on the Google Production En$\rightarrow$ Fr dataset (bold values represent best results).}
\label{tab:prodmt}
\begin{center}
\vspace{-5pt}
\scalebox{0.75}{
\begin{tabular}{l|c|c|c|c|c|c|c}
\hline \hline
Model & Eval & Eval & Test & Test & ops/timestep & Total & Training \\
      & Perplexity & BLEU & Perplexity & BLEU & & \#Parameters  & Time \\
\hline
MoE with 2048 Experts & \textbf{2.60}  & \textbf{37.27}  & \textbf{2.69} & \textbf{36.57} &85M & 8.7B & 1 day/64 k40s\\
\hline
GNMT~\citep{GNMT} & 2.78 & 35.80 & 2.87  & 35.56 & 214M & 278M & 6 days/96 k80s\\
\hline \hline
\end{tabular}
}
\end{center}
\end{table}

\paragraph{Results:} Tables \ref{tab:wmtenfr}, \ref{tab:wmtende}, and ~\ref{tab:prodmt} show the results of our largest models, compared with published results.  Our approach achieved BLEU scores of 40.56 and 26.03 on the  WMT'14 En$\rightarrow$Fr and En$\rightarrow$De benchmarks. As our models did not use RL refinement, these results constitute significant gains of 1.34 and 1.12 BLEU score on top of the strong baselines in \citep{GNMT}. The perplexity scores are also better.\footnote{Reported perplexities relative to the tokenization used by both our models and GNMT.} On the Google Production dataset, our model achieved 1.01 higher test BLEU score even after training for only one sixth of the time.

\subsection{Multilingual Machine Translation}
\label{sec:mlmt}

\paragraph{Dataset:} \citep{Johnson16} train a single GNMT \citep{GNMT} model on a very large combined dataset of twelve language pairs.  Results are somewhat worse than those for 12 separately trained single-pair GNMT models.  This is not surprising, given that the twelve models have 12 times the capacity and twelve times the aggregate training of the one model.  We repeat this experiment with a single MoE-augmented model.  See Appendix \ref{sec:appendixmt} for details on model architecture.  We train our model on the same dataset as \citep{Johnson16} and process the same number of training examples (about 3 billion sentence pairs).  Our training time was shorter due to the lower computational budget of our model.

\paragraph{Results:} Results for the single-pair GNMT models, the multilingual GNMT model and the multilingual MoE model are given in Table \ref{tab:ml}.   The MoE model achieves 19\% lower perplexity on the dev set than the multilingual GNMT model.   On BLEU score, the MoE model significantly beats the multilingual GNMT model on 11 of the 12 language pairs (by as much as 5.84 points), and even beats the monolingual GNMT models on 8 of 12 language pairs.  The poor performance on English $\rightarrow$ Korean seems to be a result of severe overtraining, as for the rarer language pairs a small number of real examples were highly oversampled in the training corpus. 

\begin{table}[h!]
\caption{Multilingual Machine Translation  (bold values represent best results).}
\label{tab:ml}
\begin{center}
\setlength\tabcolsep{2pt}
\vspace{-5pt}
\scalebox{0.8}{
\begin{tabular}{r|c|c|c|c}
\hline
\hline
   & GNMT-Mono & GNMT-Multi & MoE-Multi & MoE-Multi vs. \\
   &  &  & & GNMT-Multi \\
\hline
Parameters  & 278M / model & 278M & 8.7B & \\
ops/timestep  & 212M & 212M & 102M &  \\
training time, hardware & various & 21 days, 96 k20s & \textbf{12 days, 64 k40s} \\
\hline
 Perplexity (dev) & & 4.14 & \textbf{3.35} & -19\% \\
 French $\rightarrow$ English Test BLEU & 36.47 & 34.40 & \textbf{37.46} & +3.06 \\
 German $\rightarrow$ English Test BLEU & 31.77 & 31.17 & \textbf{34.80} & +3.63 \\
 Japanese $\rightarrow$ English Test BLEU & 23.41 & 21.62 & \textbf{25.91} & +4.29 \\ 
 Korean $\rightarrow$ English Test BLEU & 25.42 & 22.87 & \textbf{28.71} & +5.84 \\
 Portuguese $\rightarrow$ English Test BLEU & 44.40 & 42.53 & \textbf{46.13} & +3.60 \\
 Spanish $\rightarrow$ English Test BLEU & 38.00 & 36.04 & \textbf{39.39} & +3.35 \\
 English $\rightarrow$ French Test BLEU & 35.37 & 34.00 & \textbf{36.59} & +2.59 \\
 English $\rightarrow$ German Test BLEU & \textbf{26.43} & 23.15 & 24.53 & +1.38 \\
 English $\rightarrow$ Japanese Test BLEU & \textbf{23.66} & 21.10 & 22.78 & +1.68 \\
 English $\rightarrow$ Korean Test BLEU & \textbf{19.75} & 18.41 & 16.62 & -1.79 \\
 English $\rightarrow$ Portuguese Test BLEU & \textbf{38.40} & 37.35 & 37.90 & +0.55 \\
 English $\rightarrow$ Spanish Test BLEU & 34.50 & 34.25 & \textbf{36.21} & +1.96 \\
\hline
\end{tabular}
}
\end{center}
\end{table}

\vspace{-8pt}\section{Conclusion}\label{sec:conc}

This work is the first to demonstrate major wins from conditional computation in deep networks.  We carefully identified the design considerations and challenges of conditional computing and addressed them with a combination of algorithmic and engineering solutions. While we focused on text, conditional computation may help in other domains as well, provided sufficiently large training sets. We look forward to seeing many novel implementations and applications of conditional computation in the years to come.

\subsubsection*{Acknowledgments}
We would like to thank all of the members of the Google Brain and Google Translate teams who helped us with this project, in particular Zhifeng Chen, Yonghui Wu, and Melvin Johnson.  Thanks also to our anonymous ICLR reviewers for the helpful suggestions on making this paper better.

\bibliography{iclr2017_conference}
\bibliographystyle{iclr2017_conference}

\newpage
\appendix
\section*{Appendices}
\addcontentsline{toc}{section}{Appendices}
\renewcommand{\thesubsection}{\Alph{subsection}}

\subsection{Load-Balancing Loss} \label{sec:load}

As discussed in section \ref{sec:losses}, for load-balancing purposes, we want to define an additional loss function to encourage experts to receive roughly equal numbers of training examples.  Unfortunately, the number of examples received by an expert is a discrete quantity, so it can not be used in back-propagation.  Instead, we define a smooth estimator $Load(X)$ of the number of examples assigned to each expert for a batch $X$ of inputs.  The smoothness allows us to back-propagate gradients through the estimator.  This is the purpose of the noise term in the gating function.  We define $P(x, i)$ as the probability that $G(x)_i$ is nonzero, given a new random choice of noise on element $i$, but keeping the already-sampled choices of noise on the other elements.  To compute $P(x, i)$, we note that the $G(x)_i$ is nonzero if and only if $H(x)_i$ is greater than the $k^{th}$-greatest element of $H(x)$ excluding itself. The probability works out to be:

\begin{equation}
\begin{aligned}
P(x, i) = Pr\Big( (x \cdot W_g)_i + StandardNormal() \cdot Softplus((x \cdot W_{noise})_i) \\ > kth\_excluding(H(x), k, i)\Big)
\end{aligned}
\end{equation}

Where $kth\_excluding(v, k, i)$ means the kth highest component of $v$, excluding component $i$.  Simplifying, we get:

\begin{equation}
    P(x, i) = \Phi\Big(\frac{(x \cdot W_g)_i - kth\_excluding(H(x), k, i)}{Softplus((x \cdot W_{noise})_i)}\Big)
\end{equation}

Where $\Phi$ is the CDF of the standard normal distribution.

\begin{equation}
    Load(X)_i = \sum_{x \in X}P(x, i) 
\end{equation}

We can now define the load loss to be the square of the coefficient of variation of the load vector, multiplied by a hand-tuned scaling factor $w_{load}$.

\begin{equation}\label{eq:loadloss}
L_{load}(X) = w_{load} \cdot CV(Load(X))^2
\end{equation}

\paragraph{Initial Load Imbalance:}  To avoid out-of-memory errors, we need to initialize the network in a state of approximately equal expert load (since the soft constraints need some time to work).  To accomplish this, we initialize the matrices $W_g$ and $W_{noise}$ to all zeros, which yields no signal and some noise.

\paragraph{Experiments:} We trained a set of models with identical architecture (the MoE-256 model described in Appendix \ref{sec:appendixlm1b}), using different values of $w_{importance}$ and $w_{load}$.  We trained each model for 10 epochs, then measured perplexity on the test set.  We also measured the coefficients of variation in $Importance$ and $Load$, as well as ratio of the load on the most overloaded expert to the average load.  This last value is significant for load balancing purposes on distributed hardware.  All of these metrics were averaged over several training batches.

\begin{table}[h!]
\caption{Experiments with different combinations of losses. }
\label{tab:losses}
\begin{center}
\setlength\tabcolsep{3pt}
\scalebox{0.8}{
\begin{tabular}{c|c|c|c|c|c}
$w_{importance}$ & $w_{load}$ & Test Perplexity & $CV(Importance(X))$ & $CV(Load(X))$ & $\frac{max(Load(X))}{mean(Load(X))}$ \\

\hline
0.0  &  0.0  & 39.8 & 3.04 &     3.01	 & 17.80 \\
0.2  &  0.0  & \textbf{35.6} & 0.06 &     0.17    &   1.47 \\
0.0  &  0.2  & 35.7 & 0.22 &     0.04    &   1.15 \\
0.1  &  0.1  & \textbf{35.6} & 0.06 &     0.05	 & 1.14 \\
0.01 &  0.01 & 35.7 & 0.48 &     0.11    &   1.37 \\
1.0  &  1.0  & 35.7 & 0.03 &    0.02	 & \textbf{1.07} \\
\end{tabular} 
}
\end{center}
\end{table}

\paragraph{Results:} Results are reported in Table \ref{tab:losses}.  All the combinations containing at least one the two losses led to very similar model quality, where having no loss was much worse.  Models with higher values of $w_{load}$ had lower loads on the most overloaded expert.

\subsection{Hierachical Mixture of Experts} \label{sec:hierarchical}  If the number of experts is very large, we can reduce the branching factor by using a two-level hierarchical MoE.  In a hierarchical MoE, a primary gating network chooses a sparse weighted combination of ``experts", each of which is itself a secondary mixture-of-experts with its own gating network.\footnote{ We have not found the need for deeper hierarchies.}  If the hierarchical MoE consists of $a$ groups of $b$ experts each, we denote the primary gating network by $G_{primary}$, the secondary gating networks by $(G_1, G_2 .. G_a)$, and the expert networks by $(E_{0,0}, E_{0,1} .. E_{a,b})$.   The output of the MoE is given by:

\begin{equation}\label{eq:gate_expert}
y_H = \sum_{i=1}^{a}\sum_{j=1}^{b}G_{primary}(x)_i \cdot G_i(x)_j \cdot E_{i,j}(x)
\end{equation}

Our metrics of expert utilization change to the following:

\begin{equation}
Importance_H(X)_{i,j} = \sum_{x \in X}G_{primary}(x)_i \cdot G_i(x)_j
\end{equation}

\begin{equation}
Load_H(X)_{i,j} = \frac{Load_{primary}(X)_i \cdot Load_i(X^{(i)})_j}{|X^{(i)}|}
\end{equation}

$Load_{primary}$ and $Load_i$ deonte the $Load$ functions for the primary gating network and $i^{th}$ secondary gating network respectively.  $X^{(i)}$ denotes the subset of $X$ for which $G_{primary}(x)_i > 0$.  

It would seem simpler to let $Load_H(X)_{i,j} = Load_i(X_i)_j$ , but this would not have a gradient with respect to the primary gating network, so we use the formulation above.

\subsection{1 Billion Word Language Modeling Benchmark - Experimental Details}\label{sec:appendixlm1b}

\subsubsection{8-Million-Operations-per-Timestep Models}

\paragraph{Model Architecture:}  Our model consists of five layers: a word embedding layer, a recurrent Long Short-Term Memory (LSTM) layer \citep{Hochreiter:1997:LSM,Gers:2000:LFC}, a MoE layer, a second LSTM layer, and a softmax layer.  The dimensionality of the embedding layer, the number of units in each LSTM layer, and the input and output dimensionality of the MoE layer are all equal to 512.  For every layer other than the softmax, we apply drouput \citep{ZarembaSV14} to the layer output, dropping each activation with probability $DropProb$, otherwise dividing by $(1-DropProb)$.  After dropout, the output of the previous layer is added to the layer output. This residual connection encourages gradient flow \citep{HeZRS:2015:DRL}.

\paragraph{MoE Layer Architecture:}  Each expert in the MoE layer is a feed forward network with one ReLU-activated hidden layer of size 1024 and an output layer of size 512.  Thus, each expert contains $[512 * 1024] + [1024 * 512] = 1M$ parameters.  The output of the MoE layer is passed through a sigmoid function before dropout.  We varied the number of experts between models, using ordinary MoE layers with 4, 32 and 256 experts and hierarchical MoE layers with 256, 1024 and 4096 experts.   We call the resulting models MoE-4, MoE-32, MoE-256, MoE-256-h, MoE-1024-h and MoE-4096-h.  For the hierarchical MoE layers, the first level branching factor was 16, corresponding to the number of GPUs in our cluster.  We use Noisy-Top-K Gating (see Section \ref{sec:noisytopk}) with $k=4$ for the ordinary MoE layers and $k=2$ at each level of the hierarchical MoE layers.  Thus, each example is processed by exactly 4 experts for a total of 4M ops/timestep.  The two LSTM layers contribute 2M ops/timestep each for the desired total of 8M.

\paragraph{Computationally-Matched Baselines:}

The MoE-4 model does not employ sparsity, since all 4 experts are always used.  In addition, we trained four more computationally-matched baseline models with no sparsity:

\begin{itemize}
  \item MoE-1-Wide: The MoE layer consists of a single "expert" containing one ReLU-activated hidden layer of size 4096.
  \item MoE-1-Deep: The MoE layer consists of a single "expert" containing four ReLU-activated hidden layers, each with size $1024$.
  \item 4xLSTM-512: We replace the MoE layer with two additional 512-unit LSTM layers.
  \item LSTM-2048-512: The model contains one 2048-unit LSTM layer (and no MoE).  The output of the LSTM is projected down to 512 dimensions \citep{sak2014long}.  The next timestep of the LSTM receives the projected output.  This is identical to one of the models published in \citep{RafalNoam16}.  We re-ran it to account for differences in training regimen, and obtained results very similar to the published ones.
\end{itemize}

\paragraph{Training:} The models were trained on a cluster of 16 K40 GPUs using the synchronous method described in Section~\ref{sec:performance}.  Each batch consisted of a set of sentences totaling roughly 300,000 words.  In the interest of time, we limited training to 10 epochs, (27,000 steps).  Training took 12-16 hours for all models, except for MoE-4, which took 18 hours (since all the expert computation was performed on only 4 of 16 GPUs).  We used the Adam optimizer~\citep{kingma2014adam}. The base learning rate was increased linearly for the first 1000 training steps, and decreased after that so as to be proportional to the inverse square root of the step number.  The Softmax output layer was trained efficiently using importance sampling similarly to the models in \citep{RafalNoam16}.  For each model, we performed a hyper-parmeter search to find the best dropout probability, in increments of 0.1.

To ensure balanced expert utilization we set $w_{importance}=0.1$ and $w_{load}=0.1$, as described in Section \ref{sec:losses} and Appendix \ref{sec:load}.

\paragraph{Results:} We evaluate our model using perplexity on the holdout dataset, used by~\citep{chelba2013one,RafalNoam16}. We follow the standard procedure and sum over all the words including the end of sentence symbol.  Results are reported in Table~\ref{tab:lm1bresults}.   For each model, we report the test perplexity, the computational budget, the parameter counts, the value of $DropProb$, and the computational efficiency.

\begin{table}[h!]
\caption{ Model comparison on 1 Billion Word Language Modeling Benchmark. Models marked with * are from \citep{RafalNoam16}.  }
\label{tab:lm1bresults}
\begin{center}
\setlength\tabcolsep{3pt}
\scalebox{0.8}{
\begin{tabular}{l|c|c|c|c|c|c|c}
\hline
\hline
Model & Test       & Test       & ops/timestep   &  \#Params excluding & Total      & $Drop$- & TFLOPS  \\
      & Perplexity & Perplexity & (millions)     & embed. \& softmax   & \#Params  &   $Prob$ & per GPU \\
      & 10 epochs  & (final)    &                & (millions) & (billions)  &  & (observed) \\
     \hline
Kneser-Ney 5-gram* & & 67.6 & 0.00001 & & 1.8 & & \\
LSTM-512-512* & & 54.1 & 2.4 & 2.4 & 0.8 & 0.1 & \\
LSTM-1024-512* & & 48.2 & 4.7 & 4.7 & 0.8 & 0.1 & \\
\hline
LSTM-2048-512* & 45.0 & 43.7 & 9.4 & 9.4 & 0.8 & 0.1 & 0.61\\
LSTM-2048-512  & 44.7 &      & 9.4 & 9.4 & 0.8 & 0.1  & 1.21\\
4xLSTM-512     & 46.0 &      & 8.4 & 8.4 & 0.8  & 0.1 & 1.07\\
MoE-1-Wide     & 46.1 &      & 8.4 & 8.4 & 0.8  & 0.1 & 1.29\\
MoE-1-Deep     & 45.7 &      & 8.4 & 8.4 & 0.8  & 0.1 & 1.29\\ 
\hline
MoE-4          & 45.0 &      & 8.4 & 8.4 & 0.8  & 0.1 & 0.52\\
MoE-32         & 39.7 &      & 8.4 & 37.8 & 0.9 & 0.1 & 0.87\\
MoE-256        & 35.7 &      & 8.6 & 272.9 & 1.1  & 0.1 & 0.81\\
MoE-256-h      & 36.0 &      & 8.4 & 272.9 & 1.1 & 0.1 & 0.89\\
MoE-1024-h     & 34.6 &      & 8.5 & 1079.0 & 1.9 & 0.2 & 0.90\\
MoE-4096-h     & 34.1 &      & 8.9 & 4303.4 & 5.1 & 0.2 & 0.74\\
\hline
2xLSTM-8192-1024* & 34.7 & 30.6 & 151.0 & 151.0  & 1.8 & 0.25 & 1.09\\
MoE-34M           & 31.3 &      & 33.8  & 4313.9 & 6.0 & 0.3  & 1.22\\
MoE-143M          & \textbf{28.0} &      & 142.7 & 4371.1 & 6.0 & 0.4  & \textbf{1.56} \\
\hline
\hline
\end{tabular} 
}
\end{center}
\end{table}

\subsubsection{More Expensive Models}\label{sec:expensive}

We ran two additional models (MoE-34M and MoE-143M) to investigate the effects of adding more computation in the presence of a large MoE layer.  These models have computation budgets of 34M and 143M ops/timestep.   Similar to the models above, these models use a MoE layer between two LSTM layers.  The dimensionality of the embedding layer, and the input and output dimensionality of the MoE layer are set to 1024 instead of 512.  For MoE-34M, the LSTM layers have 1024 units.  For MoE-143M, the LSTM layers have 4096 units and an output projection of size 1024  \citep{sak2014long}.  MoE-34M uses a hierarchical MoE layer with 1024 experts, each with a hidden layer of size 2048.   MoE-143M uses a hierarchical MoE layer with 256 experts, each with a hidden layer of size 8192.  Both models have 4B parameters in the MoE layers.  We searched for the best $DropProb$ for each model, and trained each model for 10 epochs.

The two models achieved test perplexity of $31.3$ and $28.0$ respectively, showing that even in the presence of a large MoE, more computation is still useful.  Results are reported at the bottom of Table~\ref{tab:lm1bresults}.   The larger of the two models has a similar computational budget to the best published model from the literature, and training times are similar.  Comparing after 10 epochs, our model has a lower test perplexity by $18\%$.

\subsection{100 Billion Word Google News Corpus - Experimental Details}\label{sec:appendixgn11}

\paragraph{Model Architecture:}  The models are similar in structure to the 8-million-operations-per-timestep models described in the previous section.   We vary the number of experts between models, using an ordinary MoE layer with 32 experts and hierarchical MoE layers with 256, 1024, 4096, 16384, 65536 and 131072 experts.   For the hierarchical MoE layers, the first level branching factors are 32, 32, 64, 128, 256 and 256, respectively.

\paragraph{Training:}  Models are trained on a cluster of 32 Tesla K40 GPUs, except for the last two models, which are trained on clusters of 64 and 128 GPUs so as to have enough memory for all the parameters.   For all models, training batch sizes are approximately 2.5 million words.  Models are trained once-through over about 100 billion words.

We implement several memory optimizations in order to fit up to 1 billion parameters per GPU.   First, we do not store the activations of the hidden layers of the experts, but instead recompute them on the backwards pass.  Secondly, we modify the optimizer on the expert parameters to require less auxiliary storage:

The Adam optimizer \citep{kingma2014adam} keeps first and second moment estimates of the per-parameter gradients.  This triples the required memory.  To avoid keeping a first-moment estimator, we set $\beta_1=0$.  To reduce the size of the second moment estimator, we replace it with a factored approximation.  For a matrix of parameters, instead of maintaining a full matrix of second-moment estimators, we maintain vectors of row-wise and column-wise averages of that matrix.   At each step, the matrix of estimators is taken to be the outer product of those two vectors divided by the mean of either one.   This technique could similarly be applied to Adagrad \citep{duchi10}.

\begin{table}[h!]
\caption{ Model comparison on 100 Billion Word Google News Dataset}
\label{tab:gn11results}
\begin{center}
\setlength\tabcolsep{3pt}
\scalebox{0.8}{
\begin{tabular}{l|c|c|c|c|c|c}
\hline \hline
Model & Test       & Test       & ops/timestep   &  \#Params excluding & Total     & TFLOPS  \\
      & Perplexity & Perplexity & (millions)     & embed. \& softmax   & \#Params  & per GPU \\
      & .1 epochs  & 1 epoch    &                & (millions) & (billions)  & (observed) \\
     \hline
Kneser-Ney 5-gram & 67.1 & 45.3 & 0.00001 & & 76.0 & \\
4xLSTM-512     & 54.5 & 47.0 & 8.4 & 8.4 & 0.1  &  \textbf{1.23}\\
MoE-32         & 48.5 & 40.4 & 8.4 & 37.8 & 0.1 & 0.83\\
MoE-256-h      & 42.8 & 35.3 & 8.4 & 272.9 & 0.4 & 1.11\\
MoE-1024-h     & 40.3 & 32.7 & 8.5 & 1079.0 & 1.2 & 1.14\\
MoE-4096-h     & 38.9 & 30.9 & 8.6 & 4303.4 & 4.4 & 1.07\\
MoE-16384-h    & \textbf{38.2} & 29.7 & 8.8 & 17201.0 & 17.3 & 0.96\\
MoE-65536-h    & \textbf{38.2} & \textbf{28.9} & 9.2 & 68791.0 & 68.9 & 0.72\\
MoE-131072-h   & 39.8 & 29.2 & 9.7 & 137577.6 & 137.7 & 0.30\\
\hline \hline
\end{tabular} 
}
\end{center}
\end{table}

\paragraph{Results:} We evaluate our model using perplexity on a holdout dataset.   Results are reported in Table~\ref{tab:gn11results}.  Perplexity after 100 billion training words is 39\% lower for the 68-billion-parameter MoE model than for the baseline model.   It is notable that the measured computational efficiency of the largest model (0.30 TFLOPS/GPU) is very low compared to the other models.  This is likely a result of the fact that, for purposes of comparison to the other models, we did not increase the training batch size proportionally to the number of GPUs.  For comparison, we include results for a computationally matched baseline model consisting of 4 LSTMs, and for an unpruned 5-gram model with Kneser-Ney smoothing \citep{KneserNey95}.\footnote{While the original size of the corpus was 130 billion words, the neural models were trained for a maximum of 100 billion words.  The reported Kneser-Ney 5-gram models were trained over 13 billion and 130 billion words respectively, giving them a slight advantage over the other reported results.}

\subsection{Machine Translation - Experimental Details}
\label{sec:appendixmt}

\paragraph{Model Architecture for Single Language Pair MoE Models:} Our model is a modified version of the GNMT model described in~\citep{GNMT}.  To reduce computation, we decrease the number of LSTM layers in the encoder and decoder from 9 and 8 to 3 and 2 respectively.  We insert MoE layers in both the encoder (between layers 2 and 3) and the decoder (between layers 1 and 2). We use an attention mechanism between the encoder and decoder, with the first decoder LSTM receiving output from and providing input for the attention \footnote{For performance reasons, we use a slightly different attention function from the one described in~\citep{GNMT} - See Appendix \ref{sec:attention}}.  All of the layers in our model have input and output dimensionality of 512. Our LSTM layers have 2048 hidden units, with a 512-dimensional output projection.   We add residual connections around all LSTM and MoE layers to encourage gradient flow \citep{HeZRS:2015:DRL}. Similar to GNMT, to effectively deal with rare words, we used sub-word units (also known as ``wordpieces") \citep{Schuster:2012:JKVS} for inputs and outputs in our system. 

We use a shared source and target vocabulary of 32K wordpieces. We also used the same beam search technique as proposed in~\citep{GNMT}. 

We train models with different numbers of experts in the MoE layers.  In addition to a baseline model with no MoE layers, we train models with flat MoE layers containing 32 experts, and models with hierarchical MoE layers containing 512 and 2048 experts.  The flat MoE layers use $k=4$ and the hierarchical MoE models use $k=2$ at each level of the gating network. Thus, each input is processed by exactly 4 experts in each MoE layer.  Each expert in the MoE layer is a feed forward network with one hidden layer of size 2048 and ReLU activation.  Thus, each expert contains $[512 * 2048] + [2048 * 512] = 2M$ parameters.  The output of the MoE layer is passed through a sigmoid function.   We use the strictly-balanced gating function described in Appendix \ref{sec:batchwisemask}.

\paragraph{Model Architecture for Multilingual MoE Model:} We used the same model architecture as for the single-language-pair models, with the following exceptions:  We used noisy-top-k gating as described in Section \ref{sec:noisytopk}, not the scheme from Appendix \ref{sec:batchwisemask}.  The MoE layers in the encoder and decoder are non-hierarchical MoEs with $n=512$ experts, and $k=2$.  Each expert has a larger hidden layer of size $8192$.  This doubles the amount of computation in the MoE layers, raising the computational budget of the entire model from 85M to 102M ops/timestep.

\paragraph{Training:} We trained our networks using the Adam optimizer~\citep{kingma2014adam}. The base learning rate was increased linearly for the first 2000 training steps, held constant for an additional 8000 steps, and decreased after that so as to be proportional to the inverse square root of the step number.  For the single-language-pair models, similarly to \citep{GNMT}, we applied dropout \citep{ZarembaSV14} to the output of all embedding, LSTM and MoE layers, using $DropProb=0.4$.  Training was done synchronously on a cluster of up to 64 GPUs as described in section \ref{sec:performance}.  Each training batch consisted of a set of sentence pairs containing roughly 16000 words per GPU.

To ensure balanced expert utilization we set $w_{importance}=0.01$ and $w_{load}=0.01$, as described in Section \ref{sec:losses} and Appendix \ref{sec:load}.

\paragraph{Metrics:} We evaluated our models using the perplexity and the standard BLEU score metric. We reported tokenized BLEU score as computed by the multi-bleu.pl script, downloaded from the public implementation of Moses (on Github), which was also used in \citep{LuongPM:2015:EAANMT}. 

\begin{figure}[h!]
\centering
    \includegraphics[width=.45\textwidth]{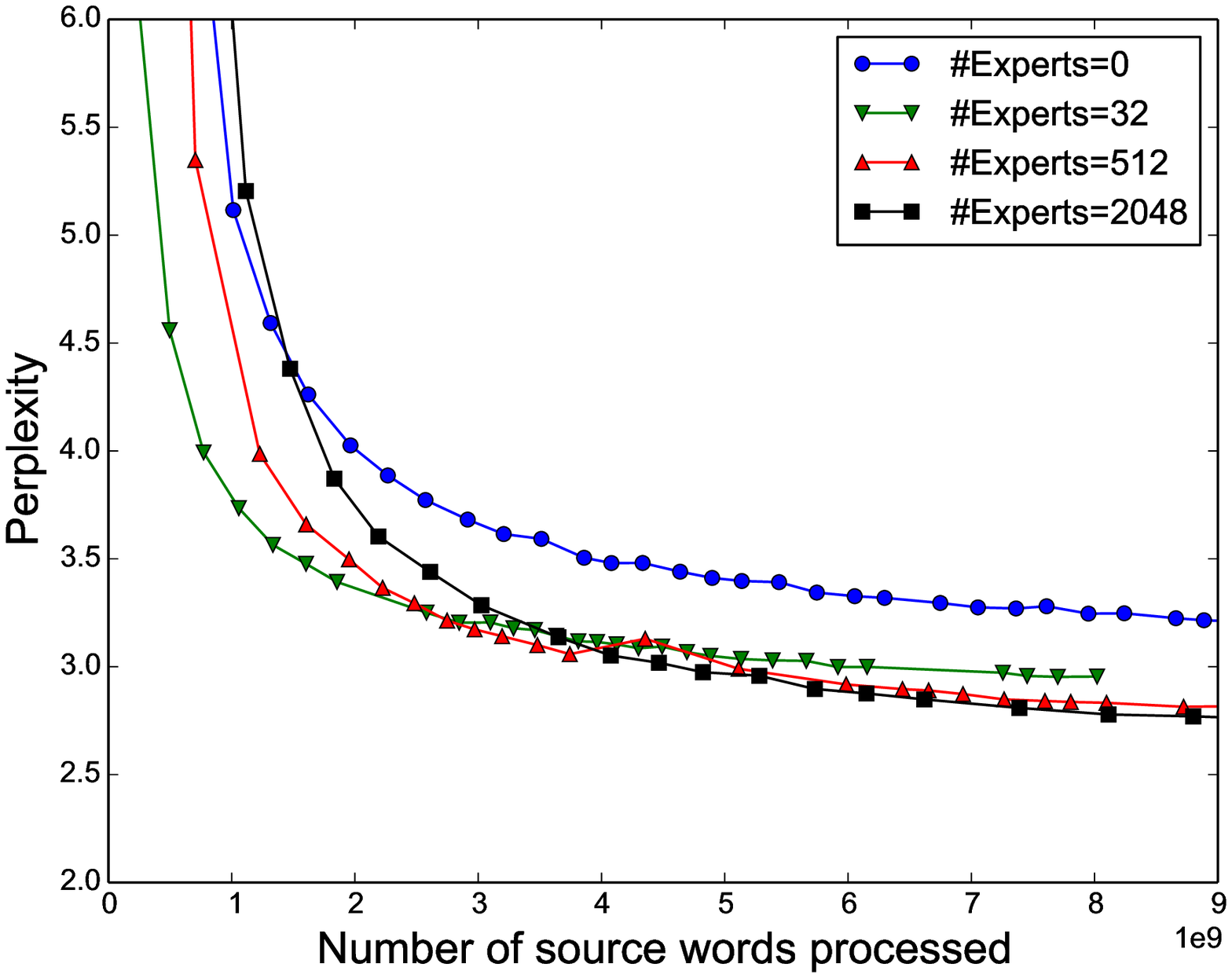}
    \includegraphics[width=.45\textwidth]{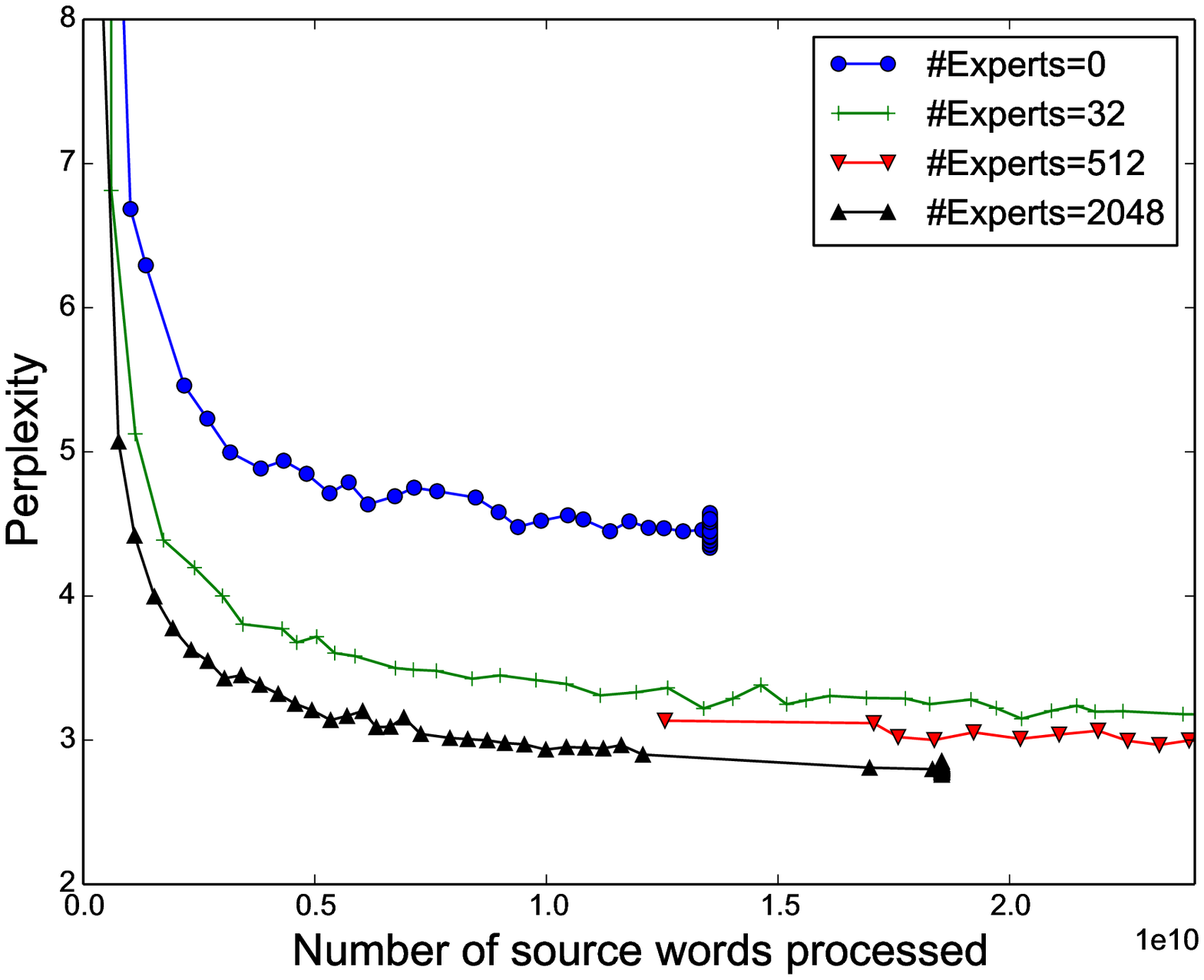}
    \caption{Perplexity on WMT'14 En$\rightarrow$ Fr (left) and Google Production En$\rightarrow$ Fr (right) datasets as a function of number of words processed.  The large differences between models at the beginning of training are due to different batch sizes.  All models incur the same computational budget (85M ops/timestep) except the one with no experts.}
    \label{fig:mt}
\end{figure}

\paragraph{Results:} Tables \ref{tab:wmtenfr}, \ref{tab:wmtende} and \ref{tab:prodmt} in Section \ref{sec:mt} show comparisons of our results to other published methods.  Figure~\ref{fig:mt} shows test perplexity as a function of number of words in the (training data's) source sentences processed for models with different numbers of experts.  As can be seen from the Figure, as we increased the number of experts to approach 2048, the test perplexity of our model continued to improve. 

We found that the experts indeed become highly specialized by syntax and/or semantics, as can be seen in Table~\ref{tab:experts}.  For example, one expert is used when the indefinite article ``a" introduces the direct object in a verb phrase indicating importance or leadership.

\begin{table}[h!]
\caption{Contexts corresponding to a few of the 2048 experts in the MoE layer in the encoder portion of the WMT'14 En$\rightarrow$ Fr translation model.  For each expert $i$, we sort the inputs in a training batch in decreasing order of $G(x)_i$, and show the words surrounding the corresponding positions in the input sentences. }
\label{tab:experts}
\centering
\label{Table}
\scalebox{0.9}{
\begin{tabular}{c|c|c}
\hline \hline
Expert 381 & Expert 752 & Expert 2004 \\
\hline
... with \textbf{researchers} ,  ... & ... plays \textbf{a} core ... & ... with \textbf{rapidly} growing ... \\
... to \textbf{innovation} . & ... plays \textbf{a} critical ... & ... under \textbf{static} conditions ... \\
... tics \textbf{researchers} . & ... provides \textbf{a} legislative ... & ... to \textbf{swift} ly ... \\
... the \textbf{generation} of  ... & ... play \textbf{a} leading ... & ... to \textbf{dras} tically ... \\
... technology \textbf{innovations} is  ... & ... assume \textbf{a} leadership ... & ... the \textbf{rapid} and ... \\
... technological \textbf{innovations} , ... & ... plays \textbf{a} central ... & ... the \textbf{fast} est ... \\
... support \textbf{innovation} throughout  ... & ... taken \textbf{a} leading ... & ... the \textbf{Quick} Method ... \\
... role \textbf{innovation} will ... & ... established \textbf{a} reconciliation ... & ... rec \textbf{urrent} ) ... \\
... research \textbf{scienti} st  ... & ... played \textbf{a} vital ... & ... provides \textbf{quick} access ... \\
... promoting \textbf{innovation} where ... & ... have \textbf{a} central ... & ... of \textbf{volatile} organic ... \\
... & ... & ... \\
\hline \hline
\end{tabular}
}
\end{table}

\subsection{Strictly Balanced Gating}\label{sec:batchwisemask} 

Due to some peculiarities in our infrastructure which have since been fixed, at the time we ran some of the machine translation experiments, our models ran faster if every expert received exactly the same batch size.  To accommodate this, we used a different gating function which we describe below.  

Recall that we define the softmax gating function to be:

\begin{equation}\label{eq:softmax}
G_\sigma(x) = Softmax(x \cdot W_g)
\end{equation}

\paragraph{Sparse Gating (alternate formulation):} To obtain a sparse gating vector, we multiply $G_\sigma(x)$ component-wise with a sparse mask $M(G_\sigma(x))$ and normalize the output. The mask itself is a function of $G_\sigma(x)$ and specifies which experts are assigned to each input example: 

\begin{equation}\label{eq:g_top_k}
G(x)_i = \frac{G_\sigma(x)_i M(G_\sigma(x))_i}{\sum_{j=1}^{n} G_\sigma(x)_j M(G_\sigma(x))_j }
\end{equation}

\paragraph{Top-K Mask:} To implement top-k gating in this formulation, we would let $M(v) = TopK(v, k)$, where:

\begin{equation}\label{eq:top_k}
TopK(v, k)_i = \begin{cases}
            1 & \text{if $v_i$ is in the top $k$ elements of $v$.} \\
            0 & \text{otherwise.}
        \end{cases}
\end{equation}

\paragraph{Batchwise Mask:} To force each expert to receive the exact same number of examples, we introduce an alternative mask function, $M_{batchwise}(X, m)$, which operates over batches of input vectors.   Instead of keeping the top $k$ values per example, we keep the top $m$ values per expert across the training batch, where $m=\frac{k|X|}{n}$, so that each example is sent to an average of $k$ experts.

\begin{equation}\label{eq:batchwisetop_k}
M_{batchwise}(X, m)_{j,i} = \begin{cases}
            1 & \text{if $X_{j,i}$ is in the top $m$ values for to expert $i$} \\
            0 & \text{otherwise}
        \end{cases}
\end{equation}

As our experiments suggest and also observed in ~\citep{DBLP:journals/corr/IoffeS15}, using a batchwise function during training (such as $M_{batchwise}$) requires modifications to the inference when we may not have a large batch of examples. Our solution to this is to train a vector $T$ of per-expert threshold values to approximate the effects of the batchwise mask.  We use the following mask at inference time:

\begin{equation}\label{eq:threshold}
M_{threshold}(x, T)_i = \begin{cases}
            1 & \text{if $x_i > T_i$} \\
            0 & \text{otherwise}
        \end{cases}
\end{equation}

To learn the threshold values, we apply an additional loss at training time which is minimized when the batchwise mask and the threshold mask are identical.

\begin{equation}\label{eq:thresholdloss}
L_{batchwise}(X, T, m) = \sum_{j = 1}^{|X|} \sum_{i=1}^n 
(M_{threshold}(x, T)_i - M_{batchwise}(X, m)_{j,i}) (X_{j, i} - T_i)
\end{equation}

\subsection{Attention Function}\label{sec:attention} 

The attention mechanism described in GNMT ~\citep{GNMT} involves a learned ``Attention Function" $A(x_i,y_j)$ which takes a ``source vector" $x_i$ and a ``target vector" $y_j$, and must be computed for every source time step $i$ and target time step $j$.  In GNMT, the attention function is implemented as a feed forward neural network with a hidden layer of size $n$.  It can be expressed as:

\begin{equation}\label{eq:gnmtattention}
A_{GNMT}(x_i, y_j) = \sum_{d=1}^{n}V_d tanh((x_iU)_d + (y_jW)_d)
\end{equation}

Where $U$ and $W$ are trainable weight matrices and $V$ is a trainable weight vector.  

For performance reasons, in our models, we used a slightly different attention function:

\begin{equation}\label{eq:ourattention}
A(x_i, y_j) = \sum_{d=1}^{n}V_d tanh((x_iU)_d) tanh((y_jW)_d)
\end{equation}

With our attention function, we can simultaneously compute the attention function on multiple source time steps and multiple target time steps using optimized matrix multiplications.  We found little difference in quality between the two functions.

\end{document}